\documentclass{article}

\usepackage{arxiv}

\usepackage[utf8]{inputenc} 
\usepackage[T1]{fontenc}    
\usepackage{hyperref}       
\usepackage{url}            
\usepackage{booktabs}       
\usepackage{amsfonts}       
\usepackage{nicefrac}       
\usepackage{microtype}      
\usepackage{lipsum}		
\usepackage{graphicx}
\usepackage{natbib}
\usepackage{doi}
\usepackage{amsmath}

\title{Do deep neural networks utilize the weight space efficiently?}

\author{ Onur Can Koyun \\
	Department of Computer Engineering\\
	Faculty of Computer and Informatics \\
	İstanbul Technical University\\
	İstanbul, Türkiye \\
	\texttt{okoyun@itu.edu.tr} \\
	\And
	Behçet Uğur Töreyin\\
    Department of Artificial Intelligence and Data Engineering \\
	Faculty of Computer and Informatics \\
	İstanbul Technical University\\
	İstanbul, Türkiye \\
	\texttt{toreyin@itu.edu.tr} \\
}

\renewcommand{\shorttitle}{\textit{arXiv} Template}

\hypersetup{
pdftitle={A template for the arxiv style},
pdfsubject={q-bio.NC, q-bio.QM},
pdfauthor={David S.~Hippocampus, Elias D.~Striatum},
pdfkeywords={First keyword, Second keyword, More},
}

\begin{document}
\maketitle

\begin{abstract}
	Deep learning models like Transformers and Convolutional Neural Networks (CNNs) have revolutionized various domains, but their parameter-intensive nature hampers deployment in resource-constrained settings. In this paper, we introduce a novel concept utilizes column space and row space of weight matrices, which allows for a substantial reduction in model parameters without compromising performance. Leveraging this paradigm, we achieve parameter-efficient deep learning models.. Our approach applies to both Bottleneck and Attention layers, effectively halving the parameters while incurring only minor performance degradation. Extensive experiments conducted on the ImageNet dataset with ViT and ResNet50 demonstrate the effectiveness of our method, showcasing competitive performance when compared to traditional models. This approach not only addresses the pressing demand for parameter efficient deep learning solutions but also holds great promise for practical deployment in real-world scenarios.
\end{abstract}


\section{Introduction}
\label{sec:intro}

Deep learning models, particularly Transformers and Convolutional Neural Networks (CNNs), have emerged as powerful tools in the field of artificial intelligence, catalyzing advancements in various domains such as computer vision, natural language processing, and speech recognition \cite{vaswani2017attention, krizhevsky2012imagenet}. However, these models often require a significant number of parameters, leading to challenges in deploying them in resource-constrained environments \cite{tan2019efficientnet, brown2020language}.

CNNs, since their resurgence with AlexNet in 2012, have been the cornerstone of computer vision tasks, offering state-of-the-art performance in areas like image classification, object detection, and more \cite{krizhevsky2012imagenet, he2016deep}. The evolution of CNN architectures, from AlexNet to more complex designs like ResNet and EfficientNet, has consistently sought to balance performance with computational efficiency \cite{he2016deep, tan2019efficientnet}.

On the other hand, Transformers, introduced by Vaswani et al. \cite{vaswani2017attention}, have revolutionized the field of natural language processing and more recently have made significant inroads into computer vision \cite{dosovitskiy2020image}. The Transformer's attention mechanism allows models to weigh the importance of different parts of the input data, a feature that has proven highly effective in various tasks. However, this capability comes at the cost of increased model complexity and a higher number of parameters. The evolution of Vision Transformers (ViTs) has led to a diverse range of models that further expand their applicability in computer vision. Notable among these is the original Vision Transformer (ViT), which applies the Transformer architecture directly to image patches for classification \cite{dosovitskiy2020image}. Following ViT, the Swin Transformer introduces a hierarchical structure using shifted windows, enhancing performance in tasks like object detection and semantic segmentation \cite{liu2021swin}.

The DETR model integrates Transformers for end-to-end object detection, showcasing the flexibility of Transformer models in adapting to different tasks \cite{carion2020end}. Additionally, models like DeiT and DINO have contributed to the field by focusing on data-efficient training and self-supervised learning, respectively \cite{touvron2021training, caron2021emerging}. Other significant contributions include Deformable DETR for more flexible object detection \cite{zhu2020deformable}, CCT for compactness in Transformer models \cite{hassani2021escaping}, and NesT which proposes a nested hierarchical approach \cite{zhang2022nested}.

Moreover, models like PVT have been developed to provide dense predictions without relying on convolutions, indicating the potential of Transformers to replace traditional CNNs in certain applications \cite{wang2021pyramid}. The T2T-ViT model takes a unique approach by converting tokens to tokens, which is a novel way of training Vision Transformers from scratch \cite{yuan2021tokens}. Additionally, the introduction of convolutions to Vision Transformers in the CvT model demonstrates the ongoing integration and evolution of Transformer and CNN architectures \cite{wu2021cvt}.

The challenge of deploying these large models in resource-limited settings, such as mobile devices or edge computing platforms, has spurred research into parameter-efficient architectures \cite{howard2017mobilenets, sandler2018mobilenetv2}. Techniques like network pruning, quantization, and knowledge distillation have been explored to reduce the size and computational requirements of these models without significantly compromising their performance \cite{han2015deep, polino2018model}.

In this context, we propose a novel approach to achieve parameter efficiency in deep learning models, particularly focusing on Transformers and CNNs. By utilizing both column space and row space of weight matrices, we present a method to significantly reduce the number of parameters in both Bottleneck and Attention layers. This technique is broadly applicable across various models in current literature. This reduction is achieved without substantial loss in performance, as demonstrated through extensive experiments on the ImageNet-1k \cite{deng2009imagenet} dataset. Our approach addresses the  need for parameter efficient deep learning models in scenarios where memory resources are scarce. The implications of this research extend beyond mere academic interest, offering a practical solution for deploying advanced AI models in real-world applications with limited resources.

\section{Related Work}
\label{sec:related}

\subsection{Efficient Deep Learning Models}
Efficient deep learning models aim to reduce computational complexity, memory requirements, and energy consumption, making them suitable for deployment in resource-constrained environments. The principal strategies include network pruning, knowledge distillation, and low-rank factorization of weight matrices.

\textit{Network Pruning:} Building upon the foundational work by Han et al. in 2015 \cite{han2015deep}, which demonstrated a significant reduction in neural network parameters with minimal accuracy loss, several studies have furthered the field of neural network efficiency and compression. The "deep compression" technique introduced by Han et al. \cite{han2015deep}, combines pruning, quantization, and Huffman coding to reduce neural network storage requirements effectively. This method involves pruning redundant connections, quantizing weights, and applying Huffman coding, substantially decreasing the storage and computational needs without sacrificing accuracy. Subsequent research has further refined these techniques. Magnitude-based weight pruning methods \cite{liang2021pruning} have proven effective in compressing networks, and complementary strategies like quantization and low-rank matrix factorization \cite{liang2021pruning} have been used alongside pruning for maximal compression. 
\textit{Knowledge Distillation:} 
First introduced by Hinton et al. \cite{hinton2015distilling}, is a method for training compact and efficient models while maintaining performance comparable to larger, more computationally demanding models. This process involves transferring knowledge from a large, complex 'teacher' model to a smaller, more efficient 'student' model. The student model is trained not only to accurately predict labels but also to emulate the label distribution produced by the teacher model. This approach allows the student model to achieve performance similar to the teacher model while being more suitable for real-time applications and devices with limited computing power.

\textit{Low-Rank Factorization:} This method decomposes weight matrices into lower rank approximations, reducing the number of parameters. Jaderberg et al. (2014) effectively applied this to CNNs for speed improvements \cite{jaderberg2014speeding}.

\subsection{Transformers}
\begin{figure*}[ht!]
\begin{center}
   \includegraphics[width=1\linewidth]{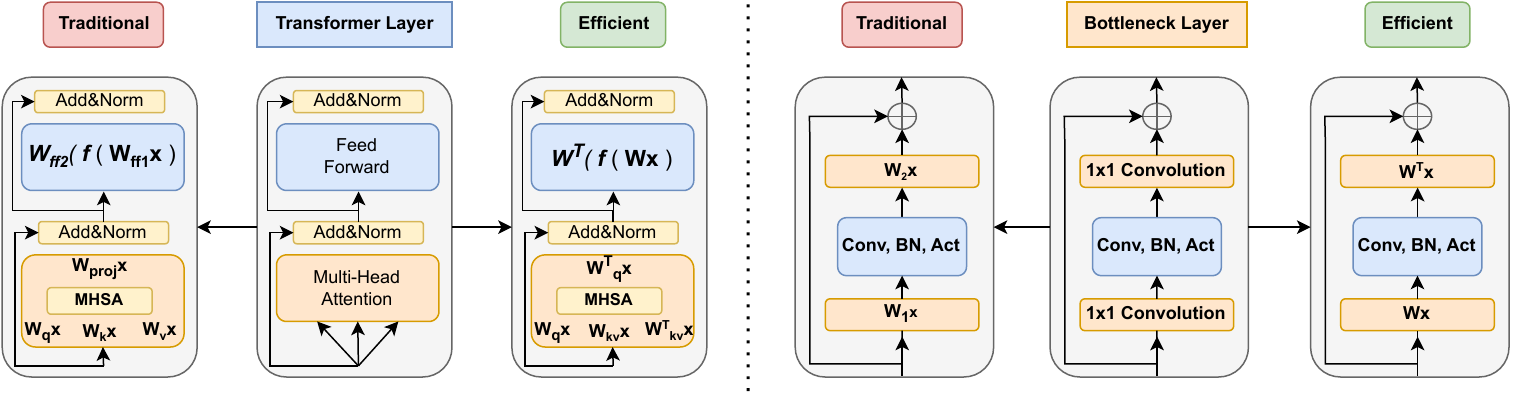}
\end{center}
   \caption{Comparison between conventional and parameter-efficient layers.
    }
\label{fig:m2}
\end{figure*}
Transformers, initially introduced by Vaswani et al. (2017) for natural language processing tasks, have been adapted for various applications including image recognition \cite{vaswani2017attention}. The self-attention mechanism allows them to model long-range dependencies efficiently. However, transformers are often parameter-intensive, which can be a drawback for deployment in memory-constrained environments.

\textit{Efficient Transformers:} A major challenge with Transformers is their high computational and memory costs, particularly in scenarios with long sequences or large inputs \cite{lin2022survey}. This has catalyzed a wave of research dedicated to enhancing the efficiency of Transformers. For instance, numerous variants such as Reformer, Linformer \cite{wang2020linformer}, and Performer have been proposed, targeting improvements in computational and memory efficiency.

A key innovation in efficient Transformer design is the development of models that address the quadratic complexity of self-attention, a central feature of the Transformer architecture. By focusing on both memory and computational efficiency, these models are more suitable for applications with constrained computational resources \cite{lin2022survey}. In the context of natural language processing, the Primer model represents a significant advancement, achieving higher efficiency in auto-regressive language modeling through architectural changes such as squaring ReLU activations and adding depthwise convolution layers \cite{so2021primer}.

The application of efficient Transformers extends beyond language tasks to fields such as computer vision. The SegFormer \cite{xie2021segformer}, for instance, integrates a novel Transformer encoder with a lightweight MLP decoder for semantic segmentation, demonstrating significant improvements in model size, runtime, and accuracy on standard datasets like ADE20K\cite{zhou2017scene} and Cityscapes \cite{cordts2016cityscapes}. This approach not only consolidates the Transformer's robustness and accuracy but also ensures efficiency, a critical aspect for real-world deployment.

\subsection{Convolutional Neural Networks (CNNs)}
Convolutional Neural Networks are predominantly used in image processing and computer vision tasks. Their layered structure allows for effective feature extraction and pattern recognition in images. LeCun et al. (1998) were pioneers in demonstrating the effectiveness of CNNs in digit recognition \cite{lecun1989handwritten}. CNNs have experienced considerable development from their early days, marked by substantial advancements in different areas. Architectural innovations have been pivotal, with the creation of AlexNet \cite{krizhevsky2012imagenet}, MobileNet \cite{howard2017mobilenets}, ResNet \cite{he2016deep}, EfficientNet \cite{tan2019efficientnet}, and GhostNet \cite{han2020ghostnet}, each representing significant breakthroughs in network design and operational efficiency.

\textit{Parameter Efficiency in CNNs:} The recent advancements in Convolutional Neural Networks (CNNs), particularly in terms of parameter efficiency, have been significantly influenced by developments in depthwise separable convolutions. This approach, exemplified in MobileNets \cite{howard2017mobilenets}., substantially reduces the number of parameters while maintaining competitive performance. 

Extending this notion, EfficientNet  \cite{tan2019efficientnet} represents a groundbreaking stride in CNN design. It employs a compound scaling method that uniformly scales network dimensions (depth, width, and resolution) using a fixed coefficient, leading to considerable gains in efficiency and performance. EfficientNet models, especially EfficientNet-B7, have achieved state-of-the-art results on benchmarks like ImageNet, CIFAR-100, and Flowers, surpassing previous models in accuracy while being significantly smaller and faster. The EfficientNet architecture, starting from the baseline model EfficientNet-B0, leverages a multi-objective neural architecture search optimizing both accuracy and computational cost (FLOPS), leading to a network with fewer parameters and higher efficiency  \cite{tan2019efficientnet}. This methodical scaling and optimization yield models that are not only compact but also computationally less demanding, allowing for higher performance with reduced resource usage.

\section{Method}

In this paper, our goal is to demonstrate that leveraging the row space and column space of  weight matrix $\mathbf{W}$ can yield comparable  performance with conventional approach. By conducting a series of experiments and analyses, we intend to showcase how these spaces contribute to the parameter efficiency of deep learning models.

Let $\mathbf{W}$ be an $m \times n$ matrix in a deep neural network layer, and $\mathbf{x}$ be a feature vector. The operation of $\mathbf{W}$ on $\mathbf{x}$ results in a linear combination of $\mathbf{W}$'s columns, where the components of $\mathbf{x}$ serve as the coefficients. This can also be interpreted as projecting $\mathbf{x}$ into the row space of $\mathbf{W}$. Therefore, the output $\mathbf{y = Wx}$ inevitably resides in the column space of $\mathbf{W}$. However, the presence of nonlinear activation functions in deep neural networks alters this characteristic. With the application of a nonlinear function $\mathcal{F}(\cdot)$, the resulting $\mathbf{y}= \mathcal{F}(\mathbf{Wx})$ no longer confines itself to the column space of $\mathbf{W}$. This change occurs as the nonlinear transformation of $\mathbf{Wx}$ shifts its dimensional orientation, moving it beyond the scope of $\mathbf{W}$'s column space. In that case, $\mathbf{W^Ty}$ does not  reside in the row space of matrix $\mathbf{W}$, implying that both the row and column spaces of matrix $\mathbf{W}$ can be independently utilized in each layer to reduce the number of parameters.

If one considers the matrix as a linear transformation from $\mathbb {R} ^{n}$ to $\mathbb {R} ^{m}$  $\mathbf{y} =\mathbf{Wx}$, then the column space of the matrix $\mathbf{W} $  equals the image of this linear transformation, similarly a linear transformation from $\mathbb {R} ^{m}$ to $\mathbb {R} ^{n}$ $\mathbf{W^Ty}$, then the row space of the matrix $\mathbf{W} $ equals the image of this transformation. Introducing nonlinearity between these projections results in a nonlinear transformation akin to that found in traditional deep neural network layers: $\mathbf{W^T} \mathcal{F} (\mathbf{Wx})$.

In our approach, we have applied this concept to transformer encoder layers in Vision Transformers (ViTs) and bottleneck layers in Residual Networks (ResNets), with a focus on reducing parameter count. Particularly, within the transformer encoder layers of ViTs, our approach effectively reduces the parameter count by half. 

\subsection{Transformer Encoder Layer}

\paragraph{Multi-Head Attention.}In MHA, $Q$,$K$ and $V$ projections are made with different weight matrices $\mathbf{W_q}$, $\mathbf{W_k}$, $\mathbf{W_v}$ and after MHA, a linear projection is made  with weight matrix $\mathbf{W_{proj}}$. Instead of using seperate weight matrices, we have utilized column and row spaces of new weight matrices  $\mathbf{W_{q}, W_{q}^T}$ and $\mathbf{W_{kv}, W_{kv}^T}$. Let $\mathbf{x}$ be a feature vector, Multi-Head Attention and linear projection can be written as;

\begin{equation}
\hat{\mathbf{x}}= MHA(\mathbf{Q,K,V}) = MHA(\mathbf{W_qx,W_{kv}x,W_{kv}^Tx}) 
\end{equation}

\begin{equation}
    Proj(\hat{\mathbf{x}},\mathbf{W_q^T}) = \mathbf{W_q^T\hat{x}}
\end{equation}

Utilizing both row and column spaces, the number of parameters in the Multi-Head Attention (MHA) layer is reduced by half.

\paragraph{Feed Forward Network.} FFN in a Transformer layer is composed of a layer normalization layer, two feed-forward layers, and a nonlinear activation function interposed between these layers. Feed-forward layers have the weight matrices $\mathbf{W_1}$ and $\mathbf{W_2}$ and bias vectors $\mathbf{b_1}$ and $\mathbf{b_2}$. Let $\mathbf{x}$ and $\mathcal{F}(\dot)$ be feature vector and nonlinear activation function, respectively. Feed-forward layer can be written as;

\begin{equation}
    FFN(\mathbf{x}) = \mathbf{W_2}\mathcal{F}(\mathbf{W_1x}+\mathbf{b_1})+\mathbf{b_2}.
\end{equation}

Instead of using seperate weight matrices $\mathbf{W_1}$ and $\mathbf{W_2}$, we leverage the column space and row space of a single weight matrix $\mathbf{W}$. This strategy leads to a reduction of the parameter count in the feed-forward layer by half. The new equation for FFN can be written as;

\begin{equation}
    FFN(\mathbf{x}) = \mathbf{W^T}\mathcal{F}(\mathbf{Wx}+\mathbf{b_1})+\mathbf{b_2}.
\end{equation}

\subsection{Bottleneck Layer}
 Introduced in the paper "Deep Residual Learning for Image Recognition" by He et al., ResNet is one of the first architectures to use bottleneck layers effectively. Architectures in the literature, leverages bottleneck layers in different ways, but with a common goal: to increase network efficiency by reducing the computational complexity while maintaining or improving model performance.

 Let $\mathcal{G}$ be a function consists of $3\times3$ convolution, normalization and nonlinear activation function. The bottleneck layer can be described as follows:

\begin{equation}
    Bottleneck(\mathbf{x}) = \mathbf{W_1^T}\mathcal{G}(\mathbf{W_2x}),
\end{equation}

where $\mathbf{W_1x}$ and $\mathbf{W_2x}$  represent $1\times1$ convolutions in bottleneck layer. Utilizing column and row spaces of weight matrix $W$, equation becomes;

\begin{equation}
    Bottleneck(\mathbf{x}) = \mathbf{W^T}\mathcal{G}(\mathbf{Wx}).
\end{equation}

Implementing our method in the Bottleneck layer reduces the number of parameters, albeit less effectively compared to its application in the Transformer layer. Consequently, we have adopted a strategy of weight sharing in each stage in ResNets to further decrease the parameter count, without adversely affecting performance. Let $\mathcal{G}_1$, $\mathcal{G}_2$, ... ,$\mathcal{G}_n$ be functions consist of $3\times3$ convolution, normalization and nonlinear activation function and $n$ is the number of bottleneck layers in a particular stage of residual network. The bottleneck layers in this stage can be written as:

\begin{equation}
    Bottleneck(\mathbf{x_1}) = \mathbf{W^T}\mathcal{G}_1(\mathbf{Wx_1}) + \mathbf{x_1},
\end{equation}
\begin{equation}
    Bottleneck(\mathbf{x_2}) = \mathbf{W^T}\mathcal{G}_2(\mathbf{Wx_2}) + \mathbf{x_2},
\end{equation}
\begin{equation}
    Bottleneck(\mathbf{x_n}) = \mathbf{W^T}\mathcal{G}_n(\mathbf{Wx_n}) + \mathbf{x_n}.
\end{equation}

By sharing weights in same stages of the network, we effectively reuse the same set of parameters for multiple operations, thereby reducing the overall parameter footprint of the model. This is particularly beneficial in deep learning architectures like ResNets, where the depth of the network can lead to a large number of parameters, potentially causing issues like overfitting and increased computational load.

Moreover, weight sharing in bottleneck layers does not significantly compromise the learning capability of the network. It allows the model to generalize better by learning reusable patterns and features across different layers. This aspect is crucial in maintaining the performance of the network while reducing its complexity.

\section{Experiment}

In this section, we detail the experimental outcomes obtained by applying our proposed method to the ImageNet-1k (IN1K) classification task \cite{deng2009imagenet}. For the implementation, the mmpretrain toolkit \cite{contributors2023openmmlab} was utilized for image classification. The IN1K dataset \cite{deng2009imagenet} comprises approximately 1.28 million training images and 50,000 validation images, distributed across 1,000 classes. For a balanced comparison, both Transformer-based and CNN-based models were trained using the training set, with the Top-1 accuracy being evaluated on the validation set. We follow the training recipe used by the
 DeiT  \cite{touvron2021training} for ViT-PE, and the same methodology was employed in Torchvision for training ResNet50.

\paragraph{ViT-PE.}
Models are trained for 300 epochs and 600 epochs. Training resolution is $224\times224$. The learning rate was scaled according to the batch size using the formula: \(\text{lrscaled} = \frac{\text{lr}}{512} \times \text{batchsize}\), following the approach of Goyal et al. \cite{goyal2017accurate,}, but modifying the base value to 512 instead of 256 as in DeiT. Results were obtained using the AdamW optimizer, employing the same learning rates as ViT \cite{dosovitskiy2020image} but with significantly reduced weight decay: $0.05$. For regularization, stochastic depth \cite{huang2016deep} was employed, which is particularly beneficial for the convergence of deep transformers \cite{fan2019reducing}. This technique was first introduced in vision transformer training by Wightman \cite{wightman2019pytorch}. We also applied regularization methods like Mixup \cite{zhang2017mixup} and Cutmix \cite{yun2019cutmix}, which contributed to improved performance.

\paragraph{ResNet50-PE.}

$224\times224$. Stochastic Gradient Descent (SGD) serves as the optimizer, with a set learning rate of 0.1 and a weight decay factor of $1e-4$. A cosine annealing schedule is used, complemented by a linear warm-up  lasting 5 epochs. During training, image augmentation techniques center cropping and flipping are employed.

\paragraph{Results.}
Table \ref{table:accuracy_comparison} demonstrates that ViT-PE, despite its lower parameter count, achieves performance on par with models such as PVTv2-B1 and DeiT-S/16. Notably, ViT-PE$^2$ outperforms PVTv2-B1, even with a reduced number of parameters. These results suggest that halving the parameter count does not significantly compromise performance, allowing for similar efficacy with less parameters.

Table \ref{tab:model_performance} indicates that the proposed method delivers performance akin to existing benchmarks. Despite sharing bottleneck layers' parameters in each stage, ResNet50-PE's performance nearly matches that of the original ResNet50 and surpasses both ResNet34 and ResNet18. Intriguingly, ResNet50-PE maintains a parameter count comparable to that of ResNet18.

\begin{table}[h]
\centering
\resizebox{0.5\textwidth}{!}{%
\begin{tabular}{|l|c|c|c|}
\hline
Method & \#Params (M) & FLOPs (G) & Top-1 Acc. (\%) \\
\hline
DeiT-T/16 \cite{touvron2021training} & 5.7 & 1.3 & 72.2 \\
PVT-T \cite{wang2021pyramid} & 13.2 & 1.9 & 75.1 \\
PVTv2-B1 \cite{wang2022pvt}& 14.0 & 2.1 & 78.7 \\
\textbf{ViT-PE$^1$ (ours)} & \textbf{11.1} & \textbf{4.6} & \textbf{77.3} \\
\textbf{ViT-PE$^2$ (ours)} & \textbf{11.1} & \textbf{4.6} & \textbf{78.8} \\
\hline \hline
DeiT-S/16 \cite{touvron2021training} & 22.1 & 4.6 & 79.9 \\
T2T-ViTt-14 \cite{yuan2021tokens} & 22.0 & 6.1 & 80.7 \\
PVT-S  \cite{wang2021pyramid} & 24.5 & 3.8 & 79.8 \\
TNT-S \cite{han2021transformer} & 23.8 & 5.2 & 81.3 \\
SWin-T \cite{liu2021swin} & 29.0 & 4.5 & 81.3 \\
CvT-13 \cite{wu2021cvt} & 20.0 & 4.5 & 81.6 \\
Twins-SVT-S \cite{chu2021twins} & 24.0 & 2.8 & 81.3 \\
FocalAtt-Tiny  \cite{yang2021focal} & 28.9 & 4.9 & 82.2 \\
PVTv2-B2  \cite{wang2022pvt} & 25.4 & 3.9 & 82.0 \\
PVTv2-B2-li  \cite{wang2022pvt} & 22.6 & 4.0 & 82.1 \\
\hline \hline
T2T-ViTt-19 \cite{yuan2021tokens} & 39.0 & 9.8 & 81.4 \\
T2T-ViTt-24 \cite{yuan2021tokens} & 64.0 & 15.0 & 82.2 \\
PVT-M \cite{wang2021pyramid} & 44.2 & 6.7 & 81.2 \\
PVT-L \cite{wang2021pyramid} & 61.4 & 9.8 & 81.7 \\
CvT-21 \cite{wu2021cvt} & 32.0 & 7.1 & 82.5 \\
TNT-B \cite{han2021transformer} & 66.0 & 14.1 & 82.8 \\
SWin-S \cite{liu2021swin} & 50.0 & 8.7 & 83.0 \\
SWin-B \cite{liu2021swin} & 88.0 & 15.4 & 83.3 \\
Twins-SVT-B \cite{chu2021twins} & 56.0 & 8.3 & 83.2 \\
Twins-SVT-L \cite{chu2021twins} & 99.2 & 14.8 & 83.7 \\
FocalAtt-Small \cite{yang2021focal} & 51.1 & 9.4 & 83.5 \\
FocalAtt-Base \cite{yang2021focal} & 89.8 & 16.4 & 83.8 \\
PVTv2-B3 \cite{wang2022pvt}& 45.2 & 6.9 & 83.2 \\
PVTv2-B4 \cite{wang2022pvt} & 62.6 & 10.1 & 83.6 \\
PVTv2-B5 \cite{wang2022pvt} & 82.0 & 11.8 & 83.8 \\
\hline
\end{tabular}%
}
\caption{Top-1 accuracy comparison in IN1K validation set on $224\times224$ resolution. 
ViT-PE$^1$ undergoes a training of 300 epochs, while ViT-PE$^2$ extends its training to 600 epochs. }
\label{table:accuracy_comparison}
\end{table}

\begin{table}[ht]
\centering
\resizebox{0.5\textwidth}{!}{
\begin{tabular}{|l|r|r|r|r|}
\hline
\textbf{Model} & \textbf{Acc@1} & \textbf{Acc@5} & \textbf{Params (M)} & \textbf{GFLOPS} \\
\hline
ResNet18 & 69.7 & 89.0 & 11.7M & 1.8 \\
ResNet34  & 73.3 & 91.4 & 21.8M & 3.66 \\
ResNet50  & 76.1 & 	92.8 & 25.6M & 4.09 \\
\textbf{ResNet50-PE}$^1$ \textbf{(Ours)} & 74.2 & 91.9 & 12.8 & 4.09 \\
\textbf{ResNet50-PE}$^2$ \textbf{(Ours)} & 75.1 & 92.8 & 13.4 & 4.09 \\
\hline
\end{tabular}
}

\caption{
The performance comparison of ResNet models on ImageNet-1K highlights two variations: ResNet50-PE$^1$ utilizes the proposed method across all its stages, whereas ResNet50-PE$^2$ applies the method exclusively to stages 3 and 4.   }
\label{tab:model_performance}
\end{table}

\section{Conclusion}

This study presents a novel approach for enhancing parameter efficiency. By exploiting the row and column spaces of the weight matrix $\mathbf{W}$, we successfully reduced the parameters of ViT and ResNet50 by half while maintaining their performance on the ImageNet dataset. The findings suggest that the strategic use of column and row spaces leads to more effective utilization of weight matrices. This method is versatile and can be implemented in any model that employs Transformer layers and bottleneck layers, significantly enhancing its applicability and potency.

\bibliographystyle{unsrtnat}
\bibliography{references}  






\end{document}